\pdfoutput=1

\documentclass[11pt]{article}

\usepackage{acl}

\usepackage{times}
\usepackage{latexsym}

\usepackage[T1]{fontenc}

\usepackage[utf8]{inputenc}

\usepackage{microtype}

\usepackage{todonotes}
\usepackage{float}
\usepackage{booktabs}
\usepackage{graphicx}
\usepackage{array}
  
\usepackage{pifont}
\newcommand{\cmark}{\ding{51}}%
\newcommand{\xmark}{\ding{55}}%

\makeatletter
\def\blfootnote{\xdef\@thefnmark{}\@footnotetext}
\makeatother

\title{Hate Speech Criteria: A Modular Approach to Task-Specific Hate Speech Definitions}

 \author{Urja Khurana$^{1*}$ \and Ivar Vermeulen$^{2}$ \and Eric Nalisnick$^{3}$ \\ \and \textbf{Marloes van Noorloos}$^{4}$ \and \textbf{Antske Fokkens}$^{1,5*}$ \\
$^1$Computational Linguistics and Text Mining Lab, Vrije Universiteit Amsterdam\\
$^2$Department of Communication Science, Vrije Universiteit Amsterdam\\
$^3$Informatics Institute, University of Amsterdam \\
$^4$Department of Criminal Law, Tilburg University \\
$^5$Dept. of Mathematics and Computerscience, Eindhoven University of Technology\\
} 

\begin{document}
\maketitle
\begin{abstract}
\textbf{Offensive Content Warning}: This paper contains offensive language only for providing examples that clarify this research and do not reflect the authors' opinions. Please be aware that these examples are offensive and may cause you distress. \\ \\
The subjectivity of recognizing \textit{hate speech} makes it a complex task. This is also reflected by different and incomplete definitions in NLP. We present \textit{hate speech} criteria, developed with perspectives from law and social science, with the aim of helping researchers create more precise definitions and annotation guidelines on five aspects: (1) target groups, (2) dominance, (3) perpetrator characteristics, (4) type of negative group reference, and the (5) type of potential consequences/effects. Definitions can be structured so that they cover a more broad or more narrow phenomenon. As such, conscious choices can be made on specifying criteria or leaving them open. We argue that the goal and exact task developers have in mind should determine how the scope of \textit{hate speech} is defined. We provide an overview of the properties of English datasets from \url{hatespeechdata.com} that may help select the most suitable dataset for a specific scenario.
\end{abstract}

\section{Introduction}
\label{sec:introduction}
The surge in online \textit{hate speech} has resulted in an increased need for its automatic detection. Its presence can be highly consequential as it creates an unsafe environment and threatens the freedom of speech \citep{kiritchenko2021confronting}. Effects of hate speech range from a personal level (e.g.\ anxiety or stress \citep{cervone2021language}) to societal level (e.g.\ discrimination or violence \citep{waldron2012harm}) and such speech can disrupt social debate severely \citep{vidgen2020directions}. Due to the large volumes of data on social media, automatizing the task is essential as hate speech can violate the law, depending on the country, in addition to its negative consequences in society. This makes automatic hate speech detection a very important task that needs to be carried out responsibly. \blfootnote{$^*$Main correspondence: \texttt{u.khurana@vu.nl} and \texttt{antske.fokkens@vu.nl}}

What is considered \textit{hate speech} is subjective \citep{fortuna-etal-2020-toxic}, there are a variety of valid viewpoints on what does (not) fall under this concept. Current hate speech datasets in NLP reflect this, having similar yet (subtly) different or incomplete definitions. For instance, similar terms are used interchangeably across publications and datasets, e.g.\ abusive, offensive, or toxic \citep{madukwe-etal-2020-data, fortuna-etal-2020-toxic}. We posit that a clear relation to (membership of) a target group of the victim sets \textit{hate speech} apart from other forms of toxic or abusive language. Underspecified definitions and guidelines increase the level of subjectivity in annotations. This subjectivity propagates into the model, which can lead to biased models \citep{sap-etal-2019-risk, davidson-etal-2019-racial}. Even if annotations are systematic, it may remain unclear which phenomena (e.g.\ target groups or types of abusive) are covered and thus captured by models.

It depends on the task for which a dataset is created whether subjectivity is desired or not. We will argue that, even for scenarios where the goal is to collect multiple viewpoints, it is important to clearly define on what aspects of the phenomenon this subjectivity is sought. At the same time, we must keep in mind that it is impossible to fully remove subjectivity when determining whether something is \textit{hate speech} or not. Even in law, where \textit{hate speech} is aimed to be defined as objectively as possible, it is inevitable that courts have difficulties interpreting such a context-dependent and sensitive topic in consistent and predictable ways \citep{van2014politicisation}. Nevertheless, a clear definition can help reduce subjectivity to borderline cases.

What would then be a good definition of \textit{hate speech}? We advocate that a good definition of \textit{hate speech} starts with a good understanding of the intention that it serves. For instance, a social media platform may want a \textit{broader} consideration of \textit{hate speech}, as it needs to keep its platform safe, in comparison to a law-enforcing model, which has to take legal action only when there is a clear punishable presence. One may decide to focus on hate speech towards one specific group or to keep this completely open to investigate which groups are considered potential targets by (crowd) annotators. 

Rather than specifying what ``the'' definition should be, we provide a meta-prescriptive setup to construct definitions and guidelines through a modular approach, where modifications can be made according to the task at hand. Concretely, we propose five criteria which should be taken into account when defining \textit{hate speech}\footnote{We emphasize that the focus of this definition is on textual \textit{hate speech}. Introducing other modalities (e.g.\ images or sound) adds other layers of complexity.} and creating annotation guidelines: (1) target groups, (2) social status of target groups, (3) perpetrator, (4) type of negative reference, (5) type of potential effect/consequence. These criteria were developed with insights from law and social science. We provide an overview of all English datasets from \textit{www.hatespeechdata.com} according to these criteria, so that people working with a specific definition in mind can easily identify which existing English datasets\footnote{We limit our overview to English for reasons of space, but plan to apply the criteria to all sets mentioned on the site in the future.} may be of direct use or provide a good starting point.

\section{Background and Motivation}
\label{sec:related_work}

In this section, we describe related work that provided the motivation and insights for the operationalization we propose in this paper.

\subsection{Annotations for Hate Speech Tasks}

The quality of annotations directly influences the quality of hate speech detection models trained on the annotated data. The subjective nature of the task makes obtaining high inter-annotator agreement, often used as a quality metric for annotations, difficult \citep{talat2017understanding}. \citet{awal2020analyzing} analyze and find evidence for inconsistency in the annotation for different widely-used hate speech datasets. They discover that some of the retweets in the dataset of \citet{founta2018large} have different labels while the tweet is the same, as also found by \citet{isaksen-gamback-2020-using}. 

Demographic factors such as language, age, and educational background have an impact on how annotation is done \citep{al-kuwatly-etal-2020-identifying,schmidt-wiegand-2017-survey} as well as expertise \citep{talat-2016-racist}. This subjectivity of the annotator also brings in possible biases of their own, as illustrated in \citet{sap-etal-2019-risk}, \citet{davidson-etal-2019-racial}, and \citet{talat-2016-racist}. \citet{sap-etal-2019-risk} show that priming annotators and making them aware of their racial bias can decrease such inclinations which could stem from misunderstanding the intent of the text.

\citet{vidgen2020directions} point out that annotators need more appropriate guidelines with clear examples to get better annotations. They also argue that it is good practice to create training sets in such a way that they address the task. We follow this point of view and start from the assumption that any annotation task (whether it is intended for training, evaluation or exploration) starts with establishing the purpose of the annotations: How will the system trained on them be used? What is investigated in case of exploratory research? 

\subsection{Task Specific Annotations}

One factor in settling how annotations can support the purpose of the task is how much subjectivity is desired. \citet{rottger2021two} distinguish two types of approaches to annotation: descriptive and prescriptive. Descriptive annotations encourage subjectivity of annotators (where inconsistency is not an issue), while prescription instructs annotators to strictly follow carefully defined criteria (the less subjectivity, the better). Accordingly, they require definitions that are either more open to interpretation or that are more specific as to what falls under the phenomenon under investigation (in our case \textit{hate speech}).

Most datasets we explored are built to serve as training data for discrete classification identifying if a message contains hate speech (or another form of abusive language). This requires consistently annotated data.\footnote{We found a notable exception in \citet{ousidhoum-etal-2019-multilingual} following a \textit{descriptive} approach, who aim to assess how people view and react to \textit{hate speech}. \citet{rottger2021two} give an overview of abusive language datasets and how they correspond to the annotation paradigm of prescriptive vs. descriptive based on the definition.} As such, most existing datasets should ideally follow a prescriptive paradigm. However, many use definitions that can introduce unintended forms of subjectivity leading to problematic forms of inconsistency. 

As mentioned in Section~\ref{sec:introduction}, even in the most prescriptive scenario of all, criminal law, experts may differ in their judgment. A certain level of disagreement is thus inevitable due to the nature of the task. This should nevertheless be limited to borderline cases and the definition should make explicit where this borderline is situated, e.g.\ should it include all potentially harmful messages to create a pleasant online environment for users or focus on the most extreme cases that potentially break the law?

While borderline cases are inevitable, there are also clear cases where there is wide agreement on a message being an example of hate speech, or contrarily benign without any signal that could possibly be problematic. In general, the fact that there is disagreement on a specific example can be valuable information \citep{aroyo2015truth}. A system trained on data that captures disagreement could for instance reflect the perception of various annotators (e.g.\ providing scores that reflect how many annotators would consider an utterance to constitute hate speech). Subjectivity is a strength in this scenario, but the racial bias reported by \citet{sap-etal-2019-risk} would still be problematic. It therefore remains desirable to raise the annotators' awareness of their biases. This would not be the case if the goal of annotating would be to investigate annotator bias rather than creating data for training or evaluating a system. Here, influence on annotators should be kept to a minimum. These examples illustrate that it is important to make conscious decisions as to where subjectivity is desired in annotations and to clearly specify which criteria annotators should not deviate from.

\subsection{Defining Hate Speech}
There is large variation in current NLP definitions and datasets. This begins with the inconsistent usage of terms. \textit{Abusive} and \textit{offensive language} are examples of terms that have been used to express the same or similar concepts \citep{schmidt-wiegand-2017-survey, talat2017understanding, fortuna-etal-2020-toxic, madukwe-etal-2020-data}.  \citet{talat2017understanding} introduce a typology that aims to further specify types of abusive language by distinguishing between (1) explicit and implicit and (2) directed or generalized forms.
We limit ourselves to \textit{hate speech}. We pose that the relation of the negativity of an abusive utterance to a target's (membership of a) specific group is a defining characteristic of \textit{hate speech}.

Even while working on the same phenomenon, there are several (subtle) differences in work addressing \textit{hate speech}.
Some datasets have a broader understanding of \textit{hate speech}, e.g.\ \citet{davidson2017automated} take a(n unintended) descriptive approach by not defining potential targets. \citet{fortuna-etal-2020-toxic} contrast the more vague definition of \citet{davidson2017automated} to the explicit list in \citet{talat2017understanding}, who intentionally focus on a more narrow phenomenon covering only racism and sexism. \citet{fortuna-etal-2020-toxic} confirm that different datasets ``provide their own flavor of hate speech" \citep[p.\ 6782]{fortuna-etal-2020-toxic}.

Varying and vague definitions can lead to inconsistencies that can (unknowingly) be problematic \citep{madukwe-etal-2020-data}. For instance, users may have a different expectation from a dataset than what its annotations actually cover. 
Ensuring that datasets are used and created appropriately starts with awareness. Therefore, we introduce \textit{hate speech criteria} (detailed in Section~\ref{sec:proposed_hs_criteria}) that can be used to construct (prescriptive or descriptive) definitions with annotational guidelines. Individual steps can be adapted depending on the task. Definitions can support a broader or more narrow focus. They can try to leave subjectivity to a minimum or explicitly keep specific aspects underspecified to collect multiple perspectives. Clear definitions can address some challenges around \textit{hate speech} identification, but not all. We elaborate on remaining open issues, such as influence of individual annotators in Section~\ref{sec:discussion}. 

Our proposal resembles prior work by \citet{kennedy2018gab}. They translate their definition into a hierarchical coding typology that is used to annotate their \textit{hate speech} dataset. They also use insights from legal (Germany, Australia, The Netherlands, and other countries), sociology, and psychology disciplines. Like us, they point out that \textit{hate speech} is treated differently per country and recognize the importance of having a negative reference relating to (membership of) a group in the utterance. 
\citet{Fortuna2018ASO} discuss differences of hate speech definitions between different sources\footnote{Platforms, code of conducts, and one scientific paper.} and recognize different dimensions that are mostly present: having a target, inciting hate or violence, to attack or diminish, and humor having a special status. \citet{zufall2020legal} introduce a schema to assess if utterances are hate speech according to the EU law. In contrast to these works, we take a broader approach and present criteria to construct definitions that fit a diverse set of operationalizations according to the desired research. Our criteria can thus support the same definitions \citet{zufall2020legal} cover, but is also wide enough to support other types of definitions. In addition, we introduce new aspects that are essential to \textit{hate speech}, such as the dominance of a group and perpetrator characteristics to our criteria. 

We furthermore go beyond these prior studies in that we provide an overview of existing English NLP datasets that address \textit{hate speech}. This overview is powered by the dimensions provided in our criteria, which are complementary to the aspects introduced in the typology by \citet{talat2017understanding}, who cover abusive language in general. Their typology does not focus on definitional \textit{hate speech} dimensions but captures the (dis)similarities between different types of abusive language. As mentioned above, they distinguish between abuse directed at an individual or generally addressing a target group and between implicit or explicit abuse. Our approach relates to that typology as follows. Their examples of generalized abusive language would typically fall under \textit{hate speech}. Speech directed at individuals also falls under \textit{hate speech} if there is direct evidence for the abuse being related to group membership. We add specifications on potential targets, group dominance, perpetrator information and the effect of the message which can encapsulate both implicit and explicit \textit{hate speech}.

\section{Proposed Hate Speech Criteria}
\label{sec:proposed_hs_criteria}

\begin{figure}[h!]
    \centering
    \includegraphics[width=0.45\textwidth]{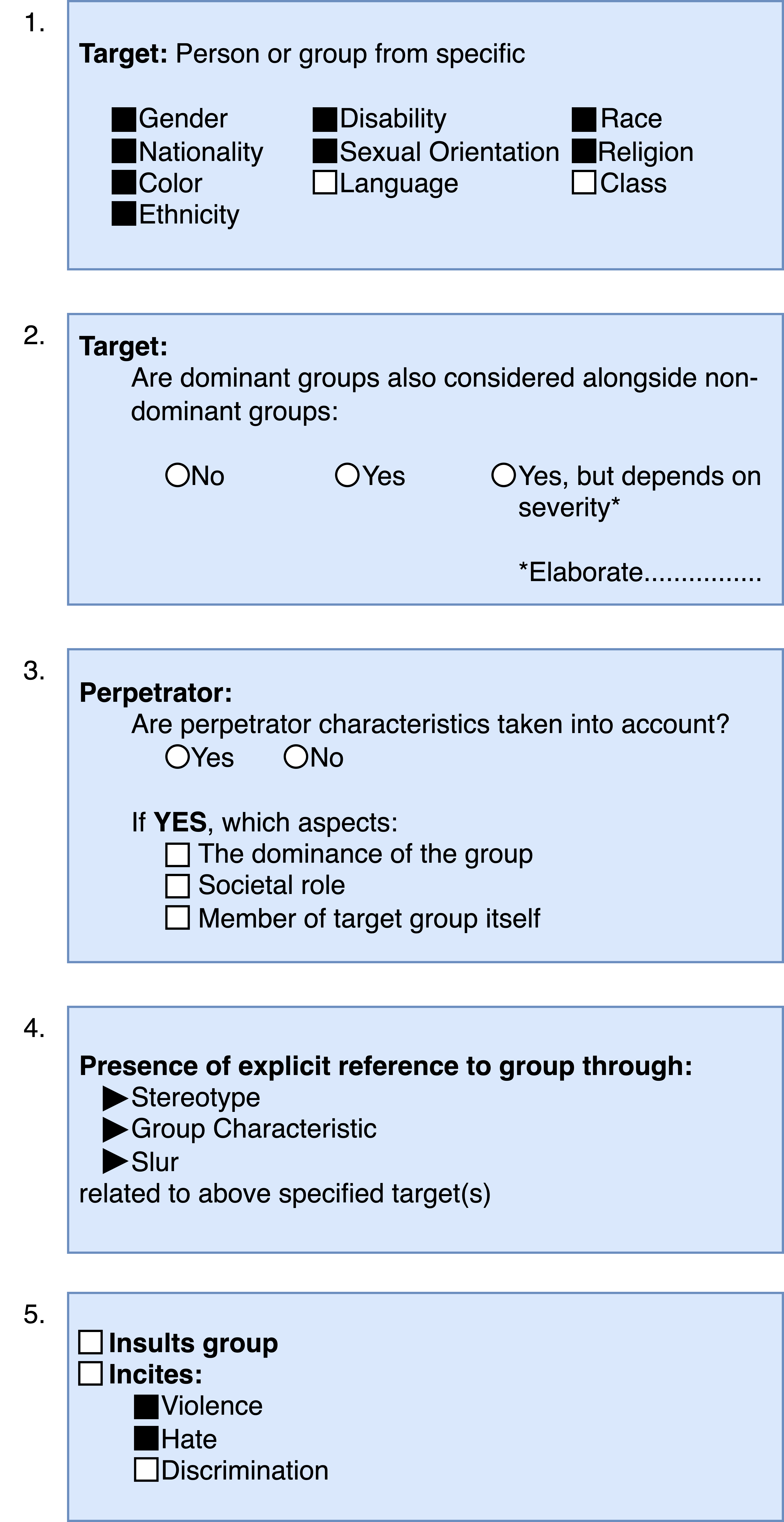}
    \caption{Our proposed \textit{Hate Speech Criteria} to support modular task-specific definition and annotation guidelines construction.}
    \label{fig:hs_criteria}
\end{figure}

This section provides our proposed procedure to define \textit{hate speech}. As outlined above, we follow the view that hate speech is characterized by problematic statements that are related to a target's (presumed) membership of a specific group. Starting from this assumption,
we propose the following criteria, represented in Figure~\ref{fig:hs_criteria}, to define the scope of \textit{hate speech}:
\begin{enumerate}
    \item Identify the target group(s).
    \item Specify the social status of the target group
    \item Consider properties of the perpetrator
    \item Identify the type of negative reference (in relation to the target) present. 
    \item Identify the potential effects/consequences of the utterance. 
\end{enumerate}

Per step, we indicate the considerations that should be taken into account. These can differ depending on the task, but certain elements are standard across many definitions found in NLP and are also generally supported by law. We call these cases \textit{standard} cases and indicate them in our figure with a filled checkbox. Other facets which are known to be considered in existing definitions (but not all) are \textit{optional} cases and these are left unfilled. This corresponds to how European countries have defined \textit{hate speech}, with some target groups being more common and/or obliged by EU law and other target groups differing among member states \citep{doi/10.2838/04029}. The options can be adjusted in any way the use case requires: one can extend the definition, narrow it down to investigate a specific form of hate speech or purposely leave a component underspecified. It is specifically possible to work with multiple definitions that apply in different legal or social contexts (e.g.\ being more lenient to what is allowed in artistic context or being more protective towards users on a social platform). Note that the criteria do not intend to distinguish different forms of \textit{hate speech}, but allow researchers to define or distinguish a specific form themselves when necessary for a task. An example of applying the criteria to a task is given in Appendix~\ref{app:example}.

\subsection{Considered Targets}

The inclusion of specific target groups depends heavily on the task (e.g.\ women- and immigrants-focused \citep{basile-etal-2019-semeval} or racism- and sexism-focused \citep{talat-hovy-2016-hateful}). For instance, a law-supporting detection system in Belgium would also consider \textit{language}\footnote{Belgian Criminal Code: Articles 377bis, 405quater, 422quater, 438bis, 442ter, 453bis, 514bis, 525bis, 532bis, and 534quater \url{https://www.ejustice.just.fgov.be/cgi_loi/change_lg.pl?language=nl&la=N&cn=1867060801&table_name=wet}} a basis of a group, while that would not be the case in the Netherlands.\footnote{Dutch Criminal Code: Articles 137d and 137e \url{https://wetten.overheid.nl/BWBR0001854/2022-03-01}} Thus, the first step in defining the scope under consideration is to specify which target groups are being considered for your task. In Figure~\ref{fig:hs_criteria}, the most common target groups are indicated as the \textbf{standard} groups. The list of possible characteristics is not exhaustive and others that historically have been disparaged can be added. Vice-versa, a study may focus on a subset of these groups. Which specific target groups people consider potential victims of hate speech can furthermore be the topic of descriptive research. In this case, it should be defined that this is intentionally left unspecified.

\subsection{Target Group Dominance} 

An important distinction that can be made in the target group is the dominance of a group in society, depending on where the model will be deployed. We define a dominant (cultural) group as a group whose members are (possibly without them being aware) positively privileged \citep{razzante2018two}, unstigmatized \citep{rosenblum2000the}, and generally favored by societal institutions \citep{marger1997chapter5}. 
Hateful sentences against non-dominant groups can be far more consequential than those addressing groups that are in power and can control the narrative. As such, objectionable speech against (an individual from) a dominant group does not necessarily have to be considered \textit{hate speech}. While for some tasks this distinction would be sensible to make, the law does not always make it, e.g.\ in The Netherlands \citep{van2014politicisation}. 

We ask the question if the task at hand also takes the dominant group into account as a potential target of \textit{hate speech}. There are three different options: The option \textit{no} excludes all forms of negative speech addressed at the dominant group from consideration. The option \textit{yes} does not distinguish between targets from the dominant group and other targets. The third option assumes that utterances targeting the dominant group can be \textit{hate speech}, but under stricter conditions. For instance, a definition may exclude the possibility of discriminating against the dominant group, but would consider calls for violence against them \textit{hate speech}.

\subsection{Speaker/Perpetrator}
The third distinction we propose is considering perpetrator characteristics \citep{geldenhuys2020demystifying}.
It should be made explicit whether, for a particular task, it matters who the perpetrator is. Because it is a common scenario in NLP that only text is available and the background of a speaker cannot be determined, there are \textbf{no standard} aspects to consider here. We describe how speaker characteristics may be taken into account for those scenarios where they can be retrieved. For instance, a person uttering possible \textit{hate speech} against their own group may be ``exempted''. It is also important to consider what such a speaker is doing with their utterance. If they are ``re-appropriating''\footnote{\textit{Reappropriate:} "to take possession for oneself that which was once possessed by another, and we use it to refer to the phenomenon whereby a stigmatized group revalues an externally imposed negative label by self-consciously referring to itself in terms of that label." - \citet{galinsky2003reappropriation}} speech to reject the negative statement \citep{galinsky2013reappropriation}, that would not be considered \textit{hate speech} while if the intention is justification, it would be. Additionally, the societal role of the perpetrator may play a role: a person in a powerful position saying something derogating can be much more harmful than an average person saying the same thing, e.g.\ a CEO of a tech company making derogatory remarks about female engineers.
In contrast, e.g.\ artists can be given more freedom due to artistic expression. Some countries, e.g.\ The Netherlands also allow more space for politicians:\footnote{\url{https://mensenrechten.nl/nl/vrijheid-van-meningsuiting}} Statements that contribute to the political debate are given more protection in lieu of freedom of expression, but remarks may not infringe other rights.\footnote{An example of a politician violating this freedom: \url{https://uitspraken.rechtspraak.nl/inziendocument?id=ECLI:NL:RBAMS:2021:7392, https://www.politico.eu/article/dutch-mep-guilty-anti-semitism-holocaust/}}

\subsection{Types of References to Target Groups} 

\textit{Hate speech} is a specific kind of abuse that is characterized by a negative reference that is either aimed at a target group or \textit{explicitly} related to membership of a target group. We thus differentiate between negative behavior toward someone from a potential target group from negative behavior \textit{because of} someone's membership of a target group. For illustration, \textit{“They should lock you up!”} clearly is a problematic message due to its threatening nature. It would nevertheless not be considered hate speech as there is no explicit reference to the individual being a part of a targeted group, even if they are in fact a member of such a group. Now, if we change it to \textit{“They should lock you up, SLUR!”} where the slur specifically targets a group, this \textit{would} be considered hate speech, as the slur clearly signals a relation between the threat and the group the target belongs to.

We explicitly state that the text should contain one (or more) of the following: (i) a stereotype (ii) a group characteristic (this can be the group itself as well) or (iii) a slur that is connected to the target groups specified in the first block. This is the only step where all the references provided are \textbf{standard} and cannot be optional. Only if the addressed task enables using more contextual clues while annotating then the reference may be found in a larger context (e.g.\ another tweet in the thread), but some evidence for the direct link between group membership and the abuse must be present. If there is no larger context, then reliance should be only on the present text. 

\subsection{Potential Consequence of Utterance} The last step in setting up the scope of \textit{hate speech} is evaluating its strength and potential effects. The actual effects need not be proved, also not in criminal law. However, the words need to be liable to incite to hatred, violence, discrimination or to insult \citep{van2014criminalising}. Most definitions consider inciting violence and hate as \textit{hate speech}, as these consequences make \textit{hate speech} stand out from other offensive expressions. These two incitements are \textbf{standard} cases. Additional broader potential consequences can also be considered, such as inciting discrimination, or a general insult toward a group. The latter is specifically recognized by Dutch law. Which possible consequences should be taken into account depends on the severity of \textit{hate speech} the task should address. It furthermore depends on the context wherein the narrative exists. Is there a relation to a threatening historical situation? Does the uttering call for exclusion of particular target groups? Furthermore, a threat can be implicitly present, i.e.\ ``What should we do with your *stereotypical object*?". While there is not an explicit \textit{call} for it, violence \textit{is} implied: destruction. The threat lies in its potential consequences. It is important to understand the implications and where the possible violence or hate stems from in a statement. Once this is understood, one may decide if different consequences, depending on severity, should apply to different targets or not. 

\section{Overview of Definitions and Datasets}
\label{sec:typology}

\begin{table*}[h!]
    \centering
    \scalebox{0.8}{
    \begin{tabular}{cccc>{\centering}p{7cm}>{\centering\arraybackslash}p{5.5cm}}
        \toprule
        & \textbf{T}  &  \textbf{ND} & \textbf{P} & \textbf{Explicit Ref} & \textbf{Effects/Consequences}  \\
        \midrule  
        \citet{talat-hovy-2016-hateful} & \cmark & \cmark & \xmark & Stereotype \& Slur & Insult, Violence, Hate, \textit{Other} \\
        \midrule
        \citet{elsherief-etal-2021-latent} & \cmark & \xmark & \xmark & Slur \& Group Characteristics & Insult, Violence, Hate, Discrimination, \textit{Other} \\
        \midrule
        \citet{kennedy2018gab} & \cmark & \xmark & \xmark & Slur, Group Characteristics, \& Stereotypes & Violence or Hate \\
        \midrule
        \citet{basile-etal-2019-semeval} & \cmark & \cmark & \xmark & \xmark & \textit{Other} \\
        \midrule
        \citet{kirk2021hatemoji} & \cmark & \xmark & \xmark & \xmark & Discrimination, \textit{Other} \\
        \midrule
        \citet{founta2018large} & \cmark & \xmark & \xmark & \xmark & Insult, Hate, \textit{Other} \\ 
        \midrule
        \citet{mandl2019overview} & \cmark & \xmark & \xmark & Stereotypes \& Group Characteristics & \xmark \\
        \midrule
        \citet{mollas2020ethos} & \cmark & \xmark & \xmark & \xmark & Insult, Violence \\
        \midrule 
        \citet{zufall2020legal} & \cmark & \xmark & \xmark & \xmark & Violence, Hate \\
        \midrule
        \citet{elsherief2018peer} & \cmark & \xmark & \xmark & \xmark &  \textit{Other} \\ 
        \midrule
        \citet{gao-huang-2017-detecting} & \cmark & \xmark & \xmark & \xmark & \textit{Other} \\
        \midrule
        \citet{qian-etal-2019-benchmark} & \cmark & \xmark & \xmark & \xmark & \textit{Other} \\
        \midrule
        \citet{ribeiro2018characterizing} & \cmark & \xmark & \xmark & \xmark & Violence, \textit{Other} \\
        \midrule
        \citet{rottger-etal-2021-hatecheck} & \cmark & \xmark & \xmark & \textit{Other} & \xmark \\
        \midrule
        \citet{chung-etal-2019-conan} & \cmark & \cmark & \xmark & \xmark & \xmark \\
        \midrule
        \citet{fanton-etal-2021-human} & \cmark & \cmark & \xmark & \xmark & \xmark \\
        \midrule
        \citet{mathew2021hatexplain} & \cmark & - & \xmark & \xmark & \xmark \\
        \midrule
        \citet{davidson2017automated} & \xmark & \xmark & \xmark & \xmark & Insult, Violence, Hate, \textit{Other} \\
        \midrule
        \citet{de-gibert-etal-2018-hate} & \xmark & \xmark & \xmark & \textit{Other} & \xmark \\ 
        \midrule
        \citet{ousidhoum-etal-2019-multilingual} & \cmark & \xmark & \xmark & \xmark & \xmark \\
        \bottomrule
    \end{tabular}}
    \caption{Overview of existing datasets according to the hate speech criteria that we propose. \textbf{T:} target groups specified, \textbf{ND:} considering only non-dominant groups specified, \textbf{Explicit Ref}: explicit reference specified, if yes; which ones, \textbf{Effects/Consequences:} effects/consequences specified, if yes; which ones. Per paper we indicate for each aspect if they are present in the definition or focus.}
    \label{tab:typology}
\end{table*}


To highlight the differences between existing datasets and to comprehend what kind of tasks they would fit, we present an overview of widely-used datasets based on their definitions. Our scope is restricted to all English datasets found on \url{hatespeechdata.com} that tackle \textit{hate speech}, since it has the most datasets and a variety of definitions.\footnote{We leave out \citet{sarkar2020chess} as it presents a challenge for hate speech detection models but does not address the task itself. We include \citet{zufall2020legal} to illustrate that our criteria also fits a legal perspective of \textit{hate speech}.} The overview can be found in Table~\ref{tab:typology}, where we indeed observe this variation.

For each step in our criteria, we indicate if it is explicitly specified in the definition or not. There is a difference between defining \textit{hate speech} and specifying a particular focus. We follow the descriptions of the annotations in the dataset for our classification. In cases where \textit{only} a definition is given, we assume that that is the focus of the dataset as well, unless stated otherwise (e.g.\ in \citet{talat-hovy-2016-hateful, basile-etal-2019-semeval}). An \xmark~ signals that the aspect is unspecified in the definition, therefore it is still possible that the specifics are present in the dataset but this cannot be guaranteed. E.g.\ several datasets might only consider non-dominant groups but do not explicitly state so, or when left unspecified it is unclear which explicit references to the group are always present.

Under column \textbf{T} we see if there are specific target groups in the definition. There are very few datasets that do not explicitly mention their target groups \citep{davidson2017automated, de-gibert-etal-2018-hate}. Although there is some overlap in groups between different datasets, there are also (subtle) differences. Due to this variety, we provide an overview of different target groups covered per dataset in Appendix~\ref{app:overview_target_groups}. 

For the second step, most definitions do not mention anything about dominance. \citet{mathew2021hatexplain} are the only ones to mention \textit{Caucasian} as a target group, which we mark with a '-' to signal that this paper is explicit about not restricting itself to non-dominant groups. Most papers with a \cmark for \textbf{ND} specifically define their targets to apply to non-dominant groups only \citep{chung-etal-2019-conan, basile-etal-2019-semeval, fanton-etal-2021-human}, with the exception of
 \citet{talat-hovy-2016-hateful}, who mention \textit{minorities}.

None of the datasets mention taking perpetrator characteristics into account (column \textbf{P}). Similarly, the explicit references are left unspecified in most datasets' definitions (column \textbf{Explicit Ref}). This means that for such datasets it cannot be guaranteed whether explicit references are present, nor whether they include specifications as to which ones. Looking at \textbf{Effects/Consequences}, \emph{violence} and \emph{hate} occur the most. Other terminology for negative relations and effects/consequences widely differs and the interpretation with respect to our criteria can be subjective (e.g.\ is `humiliate' a form of discrimination, an insult or something else?), we mark such terminology as \textit{Other}.

The idea behind the overview is that it can illustrate the need for future datasets that specify their aspects more explicitly and aids in deciding which dataset is suitable for a specific task. For instance, if a task requires a dataset that guarantees a focus on non-dominant groups, then column \textbf{ND} can easily point to the datasets that fit this prerequisite explicitly. If the dominance being specified is not very important but the presence of an explicit reference of a negative relation like a slur is, then \citet{mandl2019overview} could be fitting for the task. If, in addition, incitement of hate and violence is essential, then \citet{kennedy2018gab} should be considered. If dominance is important as well, one might consider to further annotate samples from these sets that are labeled as \textit{hate speech}, saving the time to separate `clean' messages. In combination with the overview of which datasets cover which target groups, we believe these outlines to be helpful for identifying useful datasets.

\section{Discussion}
\label{sec:discussion}

The presented \textit{hate speech} criteria aim to include those aspects of \textit{hate speech} needed to arrive at clearer definitions and to provide better annotation guidelines, while supporting a wide range of use cases. We are aware, however, that they do not provide a magic solution to all challenges around this complex phenomenon.
In this section we briefly discuss (1) possible extensions, (2) possible further specifications and (3) challenges that a clear definition cannot (fully) address on its own.

We tried to create an extensive overview of relevant aspects, but are well aware that we may have missed things. Moreover, \textit{hate speech} is strongly connected to culture and what is perceived as \emph{hate speech} may change. The criteria can thus be \textbf{extended} to cover new target groups, more perpetrator characteristics, or additional potential consequences. This particularly holds for the fourth step: \textit{Types of References to Target Groups}. We maintain that \emph{evidence} of the abuse being related to (assumed) group membership is a requirement, but researchers may decide that other clues can also serve as possible evidence for an utterance being \textit{hate speech}. These clues may include the history of a certain perpetrator, who in the past has uttered instances of \textit{hate speech} multiple times. As mentioned, the evidence may also come from e.g.\ what the utterance is responding to. 

The typology of \citet{talat2017understanding} is not included in our criteria. We explained in Section~\ref{sec:related_work} how our criteria relate to this typology. It is however straightforward to add \textbf{further specifications} to a \emph{hate speech} definition created through our criteria. Note that, even though our criteria relies on the clear presence of group characteristics, stereotypes, and/or slurs, they do not exclude implicit forms of abuse, especially since we also evaluate the potential consequences e.g.: \textit{"Everything was quite ominous with the train accident. Would like to know whether the train drivers were called StereotypicalName1, StereotypicalName2 or StereotypicalName3 \#RefugeeCrisis"} \citep{benikova2017does}. Here, the stereotypical names indicate that this falls under \textit{hate speech}. Utterances like "\textit{white revolution is the only solution}"\citep{elsherief-etal-2021-latent} may seem problematic due to the lack of an explicit slur, stereotype or target group characteristic. It nevertheless provides direct evidence, since ``white'' implies that the revolution would be against non-white and the violent nature of the threat.

A clear definition can help avoid inconsistencies and unwanted forms of subjectivity in the data, but it cannot address all challenges involved in determining whether an instance exhibits \emph{hate speech}. First, we mentioned multiple times that some form of \textbf{subjectivity} remains inevitable when dealing with \emph{hate speech}.
Though people will always differ as to where they draw the line, instructions on the level of severity that should be included with illustrating discussions can be helpful.
Even for the seemingly clear case of inciting violence, which is a core aspect, there is a vast difference between uttering \textit{"Throw tomatoes at them!"} and an actual life-threatening \textit{"Gun them down!"}. 
The class of group insult in particular can include a large variety, from merely unkind statements \textit{"Women really have a horrible sense of fashion with their white sneakers!"} to insults that question people's capabilities or attack someone's morals. Questioning a groups capabilities can lead to discrimination, especially when uttered by people with authority or in power. The remark \textit{"I'm not sexist but female comedians just are not funny!"} \citep{shvets-etal-2021-targets} may seem relatively harmless when coming from a tweeter with few followers, but when coming from an influential critic or the president of a comedians' union, it can actively harm women's careers. Attacks on a group's morals can also have an impact beyond merely insulting. For instance, saying that a non-dominant group are leeches can incite hate or lead to violent ramifications.

Explanations that illustrate the potential affect on different targets can help annotators to determine the severity of a specific statement and may help them to make more systematic decisions on where to draw the line.
This leads to the second challenge that a definition by itself cannot solve: \textbf{annotator bias}. Explanations and training may help annotators to tackle their bias and may make them more sensitive to more subtle attacks to groups they are not part of, but the affect of an annotator's background will not be completely eliminated.
Our criteria are meant to support creating a definition and guidelines. They do not mention gathering annotator information, because we believe that the definition crafted for a specific task does not change based on annotator information. However, we want to emphasize that it is essential that annotator demographics are taken into account and that the goal of the task should be kept in mind when establishing annotators' background. E.g.\ if the target group considered for the task is \textit{Gender}, it is of utmost importance to have annotators that can capture experiences of all genders. In general, it is vital to include members of potential target groups, since they are more likely to pick up on subtleties. 

A third issue that can only partially be addressed by means of a clear definition lies in the relation between \emph{hate speech} and \textbf{freedom of speech}. As mentioned, \emph{hate speech} can create unsafe environments that hamper freedom of speech. At the same time, opinions can differ regarding whether specific remarks are harmful or should be allowed because they are part of an important debate (and where marking them as \emph{hate speech} would hamper freedom of speech). Law has to clearly define what is punishable \textit{hate speech} and what on the other hand should be protected by freedom of speech. In Dutch law, a distinction is made for e.g.\ public debates where politicians are given more (but not unlimited) space for \textit{controversial} statements. Another example is that Dutch law protects members of a religion from problematic utterances (which is prohibited in all EU countries), but leaves ample space for negative statements about specific religions (decriminalized in many Western countries) \citep{van2014criminalising}. In the context of creating a safe environment for discussion, this distinction between attacking a religion or people can be hard to make sometimes and there are cases where it does not seem to make sense. The example \textit{\#BanIslam} from \citet{talat-hovy-2016-hateful}, for instance, might be aimed at religion and not at people, but it can clearly be harmful to Muslims and it is hard to see how such a hashtag would contribute to a useful public debate or discussion. In this context, both the potential harm and added value of statements to a debate should be taken into account. Though the example of \textit{\#BanIslam} seems clear, it is easy to imagine that it is not always straightforward to make this call.

The challenges mentioned above show that a clear definition may be a good starting point, but cannot solve everything. We provided examples that illustrate the complexity, but full discussions would merit individual papers on each of these topics. A final limitation we want to point out is that our overviews are currently limited to English datasets. Our framework leaves variables related to linguistic properties or cultural aspects open and can thus be easily applied to datasets covering other languages.

\section{Conclusion}
\label{conclusion}
We presented modular criteria to construct definitions and annotator guidelines to address \textit{hate speech}. These steps include aspects that have, to our knowledge, not been prominent before. We propose five components for defining hate speech: (1) identifying the target group(s), (2) specifying the consideration of dominant groups, (3) considering perpetrator characteristics, (4) finding explicit negative reference(s) of the target(s), and (5) identifying the potential consequences/effects. Based on the task at hand, the definition can be modified as, depending on the application for which \textit{hate speech} is addressed, a different description may be needed. This ties into how strictly each specific aspect needs to be defined as well: do we need annotations that are as consistent as possible (prescriptive) or do we want to investigate diversity in perspectives on this particular aspect (descriptive)?

We provided an overview of a large variety of English \textit{hate speech} datasets based on the dimensions that are present in our criteria. We hope that the criteria and discussions in this paper will motivate NLP researchers working on \textit{hate speech} to critically think about the tasks they are addressing and evaluate how fitting current definitions and datasets are for their task. The overview can then help select the most suitable datasets that can either be directly used or used as starting points that serve the task after adding further specifications. We particularly hope that the discussion in this work will help those working on new datasets to take these aspects into account from the start.

\section*{Acknowledgements}
This research was (partially) funded by the Hybrid Intelligence Center, a 10-year programme funded by the Dutch Ministry of Education, Culture and Science through the Netherlands Organisation for Scientific Research. We would additionally like to thank the reviewers for providing us with valuable feedback that has helped improving this paper.

\bibliography{anthology,custom}
\bibliographystyle{acl_natbib}

\newpage
\appendix

\begin{table*}[h!]
    \centering
    \scalebox{0.80}{
    \begin{tabular}{c>{\centering}p{2cm}>{\centering}p{2.3cm}>{\centering}p{2cm}>{\centering}p{2cm}>{\centering}p{2cm}>{\centering\arraybackslash}p{3cm}}
        \toprule
        & \textbf{Gender}  & \textbf{Origin} & \textbf{Religion} & \textbf{Sexual Orientation} & \textbf{Health} & \textbf{Other}  \\
        \midrule
        \citet{talat-hovy-2016-hateful} & Gender & Race &  &  & & \\
        \midrule
        \citet{elsherief-etal-2021-latent} & Gender (Identity), Sex & Race, Ethnicity, Nationality & Religion & Sexual Orientation & Disability, Disease & Age \\
        \midrule 
        \citet{kennedy2018gab} & Gender & Race, Ethnicity, Nationality, Regionalism & Religion, Spiritual Identity & & Mental, Physical Health & Ideology, Political Identification \\
        \midrule
        \citet{basile-etal-2019-semeval} & Women & Immigrants & & & & \\
        \midrule
        \citet{kirk2021hatemoji} & Gender & Race, Ethnicity, Nationality, Color, Descent & Religion & & & "Other identity factor" \\
        \midrule
        \citet{founta2018large} & Gender & Ethnicity, Race & Religion & Sexuality & Disability & "Attributes such as" \\ 
        \midrule 
        \citet{mandl2019overview} & Gender & Race & & Sexual Orientation & Health Condition & Political Opinion, Social Status, "or similar" \\ 
        \midrule
        \citet{mollas2020ethos} & Gender & Race, National Origin & Religion & Sexual Orientation & Disability & \\
        \midrule
        \citet{zufall2020legal} & & Race, Colour, Descent, National or Ethnic Origin & Religion & & & \\
        \midrule
        \citet{elsherief2018peer} & Gender, Sex & Race, Ethnicity, National Origin & Religion & Sexual Orientation & Disability, Disease & \\
        \midrule
        \citet{gao-huang-2017-detecting} & Gender & Ethnicity & & Sexual Orientation & & "Facet of identity" \\ 
        \midrule
        \citet{qian-etal-2019-benchmark} & Gender (Identity), Sex & Race, Ethnicity, National Origin, Caste & Religion & Sexual Orientation & Disease, Disability &  \\
        \midrule
        \citet{ribeiro2018characterizing} & Gender (Identity) & Race, Ethnicity, National Origin & Religion & Sexual Orientation & Disability, Disease & Age \\ 
        \midrule 
        \citet{rottger-etal-2021-hatecheck} & Women, Trans people & Black people, Immigrants & Muslims & Gay people & Disabled people &  \\
        \midrule
        \citet{chung-etal-2019-conan} & & & Islamophobia & & & \\ 
        \midrule
        \citet{fanton-etal-2021-human} & Women & People of Color, Romani, Migrants & Jews, Muslims & LGBT+ & Disabled people & Overweight people \\
        \midrule 
        \citet{mathew2021hatexplain} & Gender & Race, Indigenous, Refugee, Immigrant & Religion & Sexual Orientation & & \\ 
        \midrule 
        \citet{ousidhoum-etal-2019-multilingual} & Gender & Origin & Religion & Sexual Orientation & Special Needs & \\ 
        \bottomrule
    \end{tabular}}
    \caption{Overview of target groups. Each column represents a type of target, under which we indicate the specific targeted group per dataset. An empty cell indicates that the target group type was not mentioned in the definition/focus. }
    \label{tab:overview_target_groups}
\end{table*}

\section{Overview of Target Groups in Datasets}
\label{app:overview_target_groups}

Since many of the datasets have varied target groups that are taken into consideration, we present an overview of the target groups that are mentioned in their definitions, or are explicitly stated to be their focus, in Table~\ref{tab:overview_target_groups}. Due to different terminology used for related concepts, we use umbrella terms for the distinct categories and indicate the precise terms if those categories are present in the dataset (e.g. health concerns, disease, and disability grouped under health). Under \textbf{Other} we illustrate target groups that do not fit the other categories and do not occur enough to be specified by themselves. Furthermore, we also indicate if definitions keep the target groups open to unspecified ones by using wordings like "groups such as ..." (e.g. \citet{kirk2021hatemoji, mandl2019overview}). 

Datasets that do not have any target groups indicated, as can be seen in Table \ref{tab:typology}, are left out from this overview. 

When a specific focus is mentioned (e.g. \citet{chung-etal-2019-conan, basile-etal-2019-semeval}), instead of using the umbrella terms, we use the exact targets as mentioned by the paper. Moreover, when both \textit{Gender} and \textit{Gender Identity} are considered by a dataset, this is indicated as \textit{Gender (Identity)}.

\newpage

\ \\
\newpage
\section{Example of Applying the Criteria to a Task}
\label{app:example}

To showcase the utility of the proposed criteria to create a definition and associated annotation guidelines, we will apply the criteria to a prescriptive and descriptive scenario. For reasons of simplicity, we assume the questions are applied to texts that contain some form of abuse (where we use \textit{abuse} as an overarching term which includes any form of \emph{hate speech} as well as other forms potentially harmful content).

\subsection{Prescriptive scenario}

Consider the following scenario where we want to create a definition for a task that takes down \textit{hate speech} on a social media platform that goes against the Dutch law for \textit{hate speech}.\footnote{Dutch Criminal Code: Articles 137d and 137e \url{https://wetten.overheid.nl/BWBR0001854/2022-03-01}.} We initially want annotations in a prescriptive setting: as the addressed task concerns the law, we strive to reduce subjectivity to a minimum.

For each step, we fill in what is specified according to the Dutch law. Correspondingly, the target groups considered are: race, religion or philosophy of life, gender, hetero- or homo-sexuality, and physical and mental disability. As the law does not make a distinction between dominant and non-dominant groups, the task will not either. The step around perpetrator is more complex: the law does not define a distinction for the kind of perpetrator, but does allow for more lenience in a political or artistic context. We require the presence of all the explicit references mentioned in the criteria as all of them are standard cases: stereotype, group characteristic, and slur. Furthermore, the incitement of violence, hate or discrimination is considered. In addition, group insult is also seen as a consequence for all target groups except for gender.

This brings us to the following definition for the task: \textit{For this task, \textit{hate speech} is defined as language targeted at a person or group based on their race, religion or philosophy of life, gender, hetero- or homo-sexuality, and physical and mental disability and incites violence, hate or discrimination or insults a group on the basis of aforementioned targets, barring gender.}

Then, we transfer this definition using the criteria to annotation guidelines. For each step, we ask the question if the specification is present or not. If any step is not considered, we keep the lack of consideration as a note to prevent personal judgments in such cases as much as possible. If the answer to a question is \textit{yes}, the annotator can proceed to the next question. If an answer is no, the instance is not covered by the task definition of \textit{hate speech} and the annotator can directly label the instance as \textit{not hate speech}.

For this specific task, we ask the following questions to be answered for texts that contain a form of abuse:

\begin{enumerate}
    \item Does the target (group) belong to one of the following groups: \textit{race, religion or philosophy of life, gender, hetero- or homo-sexuality, and physical and mental disability}?
    \item \textbf{NOTE:} The target (group) can belong to both non-dominant and dominant groups. 
    \item Does the text contain an explicit reference to the group (related to above specified target(s)) through a stereotype, group characteristic or slur? 
    \item Does the text incite violence, hate, or discrimination or group insult? If the text incites violence, hate or discrimination, it should be labeled as \emph{hate speech}. If the text only contains a group insult proceed to the next question.
    \item Is the group insult directed at a group based on the following characteristics: \textit{race, religion or philosophy of life, hetero- or homo-sexuality, and physical and mental disability}? If yes, it should be labeled as \emph{hate speech}.
    \item If the text contains \emph{hate speech}: Is the speaker an artist or politician making the utterance in a context of their work? Please indicate ``political or artistic context'' (it can still be hate speech).
\end{enumerate}

Please note that the instructions above literally follow Dutch law and we are aware that these targets does not cover all groups (notably this is a rather limited view on LGBTQ+) and that the exclusion of gender from group insult is debatable. It should also be noted that aspects such as group dominance or who the speaker is can have an influence on a verdict, but applying those subtleties would, in this scenario, be up to judges making a final verdict rather than annotators marking potential violations of the law. For similar reasons, we choose to mark the artistic or political context as a relevant aspect rather than specifying what this would mean for the ultimate decision.

\subsection{Descriptive Scenario}

Suppose that we want to study what groups various annotators consider potential victims of \emph{hate speech}, including whether they distinguish between dominant or non-dominant groups. We focus on forms of hate speech that incite violence hate or discrimination, leaving the more vague category of group insults out. The questions they should then answer about texts containing some form of abuse are:

\begin{enumerate}
    \item Is the abuse aimed at a specific target group or a member of such a group?
    \item Does the text contain an explicit reference to the group (related to above specified target(s)) through a stereotype, group characteristic or slur?
    \item Does the text incite violence, hate, or discrimination? 
\end{enumerate}

By making it clear for which aspects the subjectivity from annotators is desired, it is easier to ensure that annotators will deviate from the given instructions for that facet and that other researchers are aware of the variation. We know, for instance, that they did check whether there is an explicit reference to the group. The awareness can help in making an informed decision if the dataset is useful for their task or not. Naturally, the outcome will remain somewhat cluttered by subjective interpretations whether a specific remark could incite, e.g., discrimination. We can however distinguish between these motivations by making annotators answer the question rather than making them label the data. As such, we learn whether their reason for `no' was related to the specific group being a potential target or to the severity or nature of the abuse.

For reasons of simplicity, we left the criterion of perpetrator information out of this example. Investigating this aspect would probably require a different setup. For instance, first defining \emph{hate speech} as something that incites violence, hate or discrimination, and then asking the following questions:

\begin{enumerate}
    \item Is the abuse aimed at a specific target group or a member of such a group?
    \item Does the text contain an explicit reference to the group (related to above specified target(s)) through a stereotype, group characteristic or slur?
    \item Is the speaker a member of the target group?
    \item Do you consider this utterance to be \emph{hate speech} based on the definition provided above?
\end{enumerate}

Note that the examples in this appendix are included for purposes of illustration as how definitions may help specifying annotation tasks. We are well aware that providing a good annotation setup, especially for descriptive scenarios, is complex. As many aspects mentioned in this paper, the next step of actually setting up such tasks can merit a paper on its own.

\end{document}